\pdfoutput=1

\PassOptionsToPackage{prologue,dvipsnames}{xcolor}

\documentclass[11pt]{article}

\usepackage[final]{acl}

\usepackage{times}
\usepackage{latexsym}

\usepackage[T1]{fontenc}

\usepackage[utf8]{inputenc}

\usepackage{microtype}

\usepackage{inconsolata}

\usepackage{graphicx}
\usepackage{multirow}
\usepackage{amsmath}
\usepackage{cleveref}

\title{Intent Matters: Enhancing AI Tutoring with \\Fine-Grained Pedagogical Intent Annotation}

\author{Kseniia Petukhova, Ekaterina Kochmar\\
  Mohamed bin Zayed University of Artificial Intelligence \\
  \texttt{\{kseniia.petukhova, ekaterina.kochmar\}@mbzuai.ac.ae}\\}

\begin{document}
\maketitle
\begin{abstract}
Large language models (LLMs) hold great promise for educational applications, particularly in intelligent tutoring systems. However, effective tutoring requires alignment with pedagogical strategies -- something current LLMs lack without task-specific adaptation. In this work, we explore whether fine-grained annotation of teacher intents can improve the quality of LLM-generated tutoring responses. We focus on \texttt{MathDial}, a dialog dataset for math instruction, and apply an automated annotation framework to re-annotate a portion of the dataset using a detailed taxonomy of eleven pedagogical intents. We then fine-tune an LLM using these new annotations and compare its performance to models trained on the original four-category taxonomy. Both automatic and qualitative evaluations show that the fine-grained model produces more pedagogically aligned and effective responses. Our findings highlight the value of intent specificity for controlled text generation in educational settings, and we release our annotated data and code to facilitate further research: \url{https://github.com/Kpetyxova/autoTree/tree/main/mathdial}

\end{abstract}

\section{Introduction}

Human tutoring is a cornerstone of educational development, playing a vital role in empowering learners and fostering societal progress. One-on-one tutoring has long been recognized as highly effective~\citep{bloom19842}; however, its widespread implementation is constrained by the limited availability of qualified tutors. Recent advancements in LLMs have shown great promise in educational contexts~\citep{wang2024large,gan2023large}, leading to the emergence of LLM-powered intelligent tutoring systems (ITS)~\citep{pal2024autotutor,liu2024teaching} and the use of LLMs as tutors via advanced prompting strategies~\citep{denny2024generative,mollick2024instructors}. These AI tutors serve a range of educational objectives~\citep{wollny2021we}, with one of the most prominent being the remediation of student mistakes and confusion -- an area that continues to drive the development of AI tutoring systems~\citep{macina2023mathdial,wang2023bridging}.

While LLMs do well both at generating human-like conversations and at addressing various reasoning tasks, such as commonsense reasoning and basic mathematical reasoning~\citep{achiam2023gpt,kojima2022large,laskar2023systematic,yang2024harnessing}, they cannot be directly deployed in educational systems without significant adaptation. Effective tutoring requires more than fluent conversation -- it involves guiding learners to discover answers on their own. Rather than simply providing solutions, a good tutor employs strategies such as giving hints, asking questions in a Socratic dialog~\citep{carey2004socratic}, and encouraging active problem-solving. As such, LLM-based tutors should ideally align with human tutoring strategies~\citep{nye2014autotutor} and active learning practices shown to enhance student outcomes~\citep{freeman2014active}.

In order to have such models, we need dialog tutoring datasets. \texttt{MathDial}~\citep{macina2023mathdial} is one such dataset, comprising tutor-student dialogs centered around math reasoning tasks. Each teacher utterance is labeled with one of four pedagogical move types from \citet{macina2023mathdial}: {\em Focus} (guiding task progress), {\em Probing} (encouraging conceptual exploration), {\em Telling} (providing help when students are stuck), or {\em Generic} (non-pedagogical conversational turns). These annotations were provided by teachers during data collection to better scaffold student learning. While this four-category taxonomy offers a helpful high-level structure, it lacks the fine-grained detail needed for advanced applications such as controlled response generation, pedagogical analysis, and behavior modeling in AI tutors. At the same time, finer-grained annotations may enable better interpretability, improved pedagogical alignment, and greater flexibility in guiding student learning experiences.

Although \texttt{MathDial}'s original taxonomy includes only four broad categories, the authors also provide an expanded set of eleven fine-grained intents, which could offer greater control and variety in AI-generated tutoring responses. Building on this, in this work, we apply a fully automated framework for conversational discourse annotation~\citep{petukhova2025fully} to {\bf re-annotate a portion of the MathDial dataset using the finer-grained eleven-intent taxonomy}. This annotation framework uses LLMs to automatically construct a decision tree from the taxonomy and use it to label utterances, providing a scalable alternative to manual annotation. This approach has demonstrated superior performance compared to crowdworkers in annotating dialog with speech functions taxonomy~\cite{eggins2004analysing}.

\textbf{Our goal in this work is to assess whether such more detailed annotations can improve the quality of LLM-based tutoring through fine-tuning models on both the original and re-annotated data.} Specifically, we fine-tune \texttt{Mistral-7B-Instruct} on the original coarse-grained as well as the new fine-grained annotation, and compare the generated tutor responses using automatic metrics and human evaluation. Our results demonstrate that \textbf{the fine-grained model produces more pedagogically aligned and effective responses}. To facilitate further research and development, we release a public repository containing both the code and the re-annotated dataset.\footnote{Available at \url{https://github.com/Kpetyxova/autoTree/tree/main/mathdial}.}

\section{Background \& Related work}

\subsection{The \texttt{MathDial} Dataset}

We build on the foundational work of \citet{macina2023mathdial}, whose dataset provide an invaluable basis for advancing pedagogically aligned dialog systems. \texttt{MathDial} is a large-scale, high-quality dialog tutoring dataset focused on multi-step math reasoning problems. Unlike previous datasets that suffer from low pedagogical quality, small size, or lack of grounding, \texttt{MathDial} provides rich annotations grounded in realistic student confusions and pedagogical strategies. The authors introduce a novel semi-synthetic data collection framework that pairs expert human teachers with LLMs simulating students and their errors, enabling scalable and controlled creation of educational dialogs that closely mimic authentic tutoring scenarios. This approach effectively addresses privacy concerns and quality issues associated with crowdsourcing or classroom recordings.

The authors' methodology consists of a Wizard-of-Oz-inspired framework~\citep{kelley1984iterative}, where expert teachers engage in one-on-one tutoring dialogs with LLMs acting as students. These student models are carefully prompted with student profiles and frequently occurring conceptual errors generated using temperature sampling over diverse reasoning paths produced by LLMs. The math word problems (MWPs) used are sourced from GSM8K~\citep{cobbe2021training}. Teachers are instructed to scaffold student understanding using a taxonomy of four pedagogical moves: {\em Focus}, {\em Probing}, {\em Telling}, and {\em Generic}, with additional fine-grained intents (see \Cref{tab:mathdial}).

\begin{table*}[htbp]
\scriptsize
\centering
\resizebox{\linewidth}{!}{
\begin{tabular}{|p{0.07\textwidth}|p{0.2\textwidth}|p{0.55\textwidth}|}
    \hline
    \textbf{Category} & \textbf{Intent} & \textbf{Example} \\
    \hline
    \multirow{3}{*}{\centering \textbf{Focus}} 
    & Seek Strategy & \textit{So what should you do next?} \\ \cline{2-3}
    & Guiding Student Focus & \textit{Can you calculate \dots ?} \\ \cline{2-3}
    & Recall Relevant Information & \textit{Can you reread the question and tell me what is \dots ?} \\
    \hline
    \multirow{4}{*}{\centering \textbf{Probing}} 
    & Asking for Explanation & \textit{Why do you think you need to add these numbers?} \\ \cline{2-3}
    & Seeking Self Correction & \textit{Are you sure you need to add here?} \\ \cline{2-3}
    & Perturbing the Question & \textit{How would things change if they had \dots items instead?} \\ \cline{2-3}
    & Seeking World Knowledge & \textit{How do you calculate the perimeter of a square?} \\
    \hline
    \multirow{2}{*}{\centering \textbf{Telling}} 
    & Revealing Strategy & \textit{You need to add \dots to \dots to get your answer.} \\ \cline{2-3}
    & Revealing Answer & \textit{No, he had \dots items.} \\
    \hline
    \multirow{2}{*}{\centering \textbf{Generic}} 
    & Greeting/Farewell & \textit{Hi \dots, how are you doing with the word problem? Good Job! Is there anything else I can help with?} \\ \cline{2-3}
    & General Inquiry & \textit{Can you go walk me through your solution?} \\
    \hline
\end{tabular}
}
\caption{Teacher moves with examples of utterances and their intents from \texttt{MathDial}~\cite{macina2023mathdial}.}
\label{tab:mathdial}
\end{table*}

Crucially, before writing a response, teachers must annotate the pedagogical move being employed, encouraging more intentional strategy use. The dialogs are also grounded in metadata, including the specific confusion, full problem, step-by-step solutions, and whether the confusion was resolved, thus offering rich signals for training AI tutors.

Empirical evaluation demonstrates that models fine-tuned on \texttt{MathDial} significantly outperform both zero-shot and instruction-tuned larger LLMs like \texttt{ChatGPT} in terms of correctness and equitable tutoring~\citep{macina2023mathdial}. Notably, fine-tuned open-source models achieved similar rates of student problem-solving success while reducing the incidence of ``telling'' -- prematurely giving away solutions. Human evaluations confirmed that these fine-tuned models were more coherent, correct, and pedagogically effective than large prompted models.

\subsection{Annotation Framework}

While manual discourse annotation is costly and time-consuming, advances in LLM-based annotation present a promising alternative with demonstrated improvements in speed, consistency, and cost-effectiveness~\citep{gilardi2023chatgpt,hao2024fullanno}. \citet{petukhova2025fully} have recently proposed an open-source pipeline for fully automated discourse annotation using LLMs. Specifically, this pipeline automates the construction of hierarchical tree annotation schemes and the annotation of utterances within dialogs, making it a promising and scalable approach for enriching the \texttt{MathDial} dataset with more detailed teacher intent annotations.

\citet{petukhova2025fully} explore multiple configurations for tree construction and annotation, including binary and non-binary structures, frequency-based grouping, and optimal split strategies, and report that the frequency-guided optimal split selection using \texttt{GPT-4o} outperforms crowdworkers on dialog annotation tasks based on the taxonomy of speech functions~\cite{eggins2004analysing}, while reducing total annotation time from over 30 hours to under 1.5 hours. Therefore, in our work, we adopt this configuration using the publicly available implementation.\footnote{\url{https://github.com/Kpetyxova/autoTree}}

\subsection{Controlled Generation}

Controlled text generation (CTG) aims to direct language models to produce outputs that adhere to specific attributes or constraints, such as sentiment, style, or intent. A prevalent method in CTG involves fine-tuning models with prompts that include explicit intent labels, enabling the generation of text aligned with desired behaviors~\citep{liang2024controllable}.

Instruction fine-tuning has emerged as an effective strategy for this purpose. By training models on datasets where prompts are augmented with natural language instructions or intent labels, models learn to condition their outputs accordingly. For instance, the InstructCTG framework demonstrates how conditioning on natural language descriptions and demonstrations of constraints allows models to generate text that satisfies various requirements without altering the decoding process~\citep{zhou2023controlled}.

This approach is particularly beneficial in educational contexts, where aligning generated content with pedagogical strategies is crucial. By fine-tuning models with prompts that specify instructional intents, AI tutors can provide more effective and tailored support to learners~\cite{jia2025fine}.

\section{Re-annotating the \texttt{MathDial} Dataset}

\subsection{Tree Creation}
\begin{figure*}[h]
  \centering
  \includegraphics[width=\linewidth]{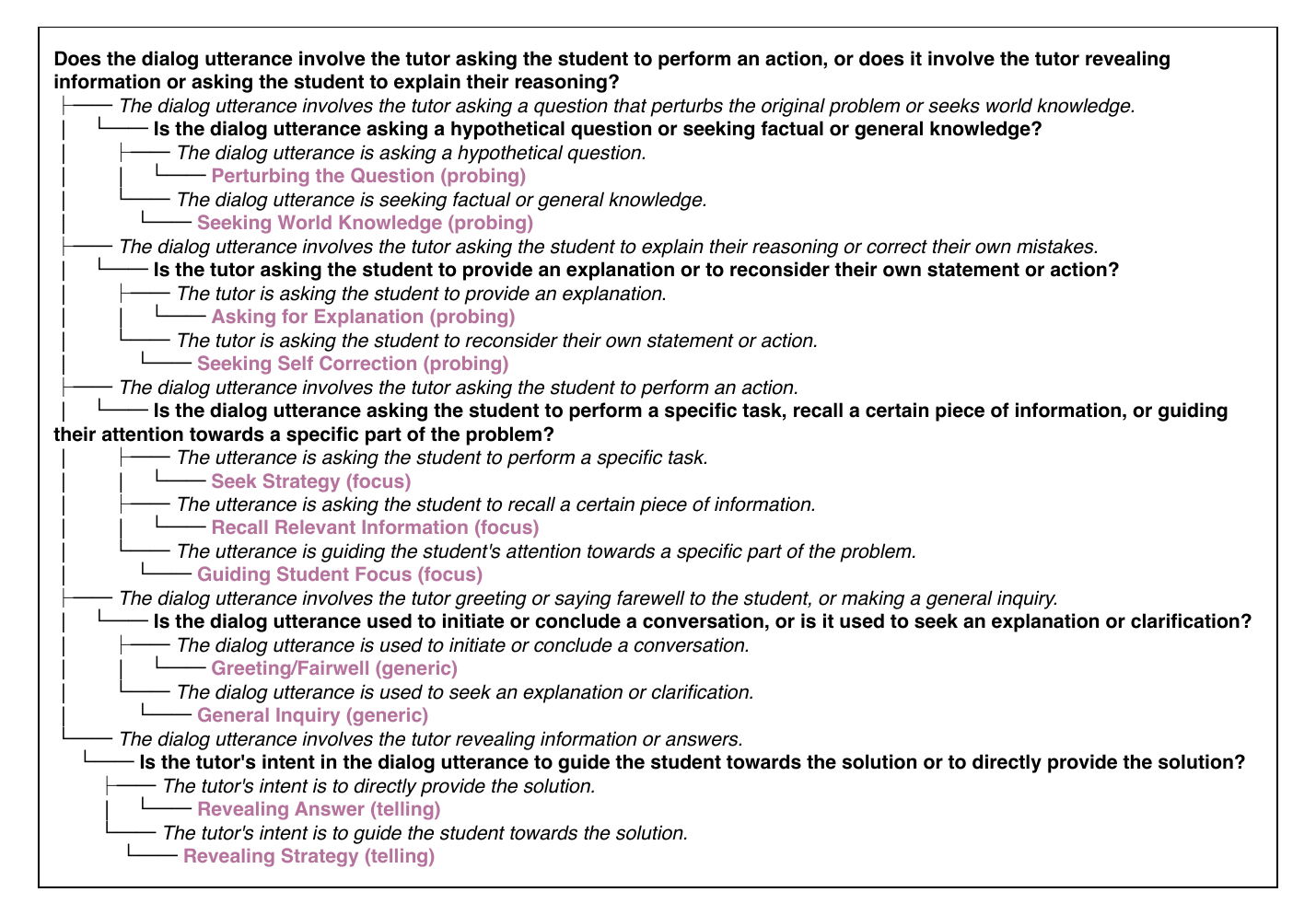}
  \caption{Tree created for the extended taxonomy of the \texttt{MathDial} dataset using the framework from \citet{petukhova2025fully}. Questions corresponding to tree nodes are in \textbf{bold}, possible answers that represent branches are in \textit{italics}, and leaf nodes, representing the eleven intents, are in \textbf{\textcolor{Orchid}{purple bold}}.}
  \label{fig:mathdial_tree}
\end{figure*}

To construct a tree for the extended taxonomy proposed in \texttt{MathDial}, we used the best framework configuration from \citet{petukhova2025fully} -- frequency-guided optimal split selection and backtracking. This method iteratively selects among the candidate splits by scoring them and choosing the highest-ranked one, with backtracking employed if a viable partition cannot be formed. Additionally, the approach biases the tree construction toward more frequent classes, making them quicker to reach and producing trees that better reflect real-world class distributions. The tree was generated based on eleven intent names and their corresponding examples (see Table \ref{tab:mathdial}). The resulting tree, presented in textual form in \Cref{fig:mathdial_tree}, has a depth of two with five branches emerging from the root node.

Interestingly, most intents are grouped according to high-level categories defined in~\citet{macina2023mathdial}, except for the {\em Probing} intents, which are split into two separate groups: (1) \textit{Asking for Explanation} and \textit{Seeking Self-Correction}, and (2) \textit{Perturbing the Question} and \textit{Seeking World Knowledge}. While this split was not predefined, it is interpretable: the first group centers on prompting students to reflect on and assess their reasoning, whereas the second group encourages them to explore broader or external concepts beyond the immediate problem.

\subsection{Annotation}

\paragraph{Data Preprocessing} Out of 2,861 dialogs, we randomly selected 500 dialogs for training, 100 for validation, and 100 for testing.

\begin{figure*}[htbp]
  \centering
  \includegraphics[width=\linewidth]{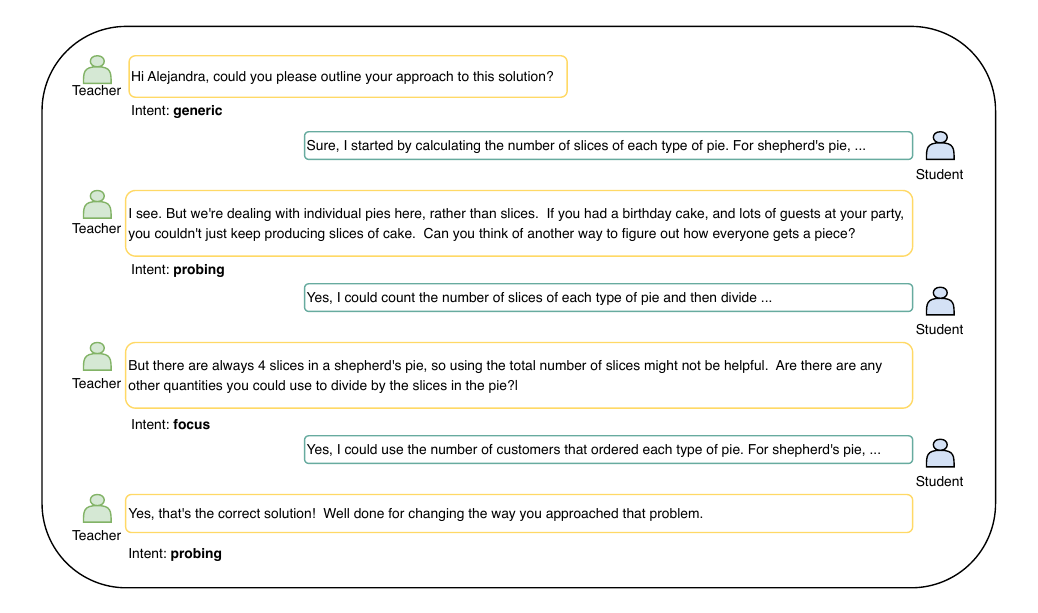}
  \caption{An example of teacher utterances and their annotated intents from \texttt{MathDial}.}
  \label{fig:annot_mathdial_example}
\end{figure*}

An example of the original tutor intent annotation in \texttt{MathDial} is shown in \Cref{fig:annot_mathdial_example}. A single label is applied to each teacher utterance in the original annotation, which, while effective for high-level analysis, may limit flexibility in downstream applications requiring finer-grained control. For instance, an utterance \textit{[I see.]$_1$ [But we're dealing with individual pies here, rather than slices.]$_2$ [If you had a birthday cake, and lots of guests at your party,
you couldn't just keep producing slices of cake.]$_3$ [Can you think of another way to figure out how everyone gets a piece?]$_4$} in {\tt MathDial} is annotated as {\em Probing}. However, this utterance comprises several discourse units with distinct functions: segment [1] appears to be \textit{Generic}, segment [2] aligns with \textit{Focus} (specifically, \textit{Guiding Student Focus}) as it redirects the student's attention, segment [3] fits the \textit{Probing} category, and segment [4] corresponds to \textit{Focus} (\textit{Seek Strategy}) because it prompts the student to think of an alternative solution.

In contrast, in other cases, the assigned label appears to follow the final part of the utterance. For example, the utterance \textit{[But there are always 4 slices in a shepherd's pie, so using the total number of slices might not be helpful.]$_1$ [Are there any other quantities you could use to divide by the slices in the pie?]$_2$} is labeled as {\em Focus}. Here, while {\em Focus} applies to the second sentence [2], it would be more appropriate to label the first sentence [1] as {\em Telling}. This inconsistency -- where labels are sometimes based on the first segment and other times on the last -- underscores the potential benefits of a more fine-grained and consistent annotation approach for certain downstream tasks.

Ideally, annotation should be performed at the elementary discourse units (EDUs) level rather than entire utterances. EDUs are segments of text that typically correspond to clauses~\cite{jurafskyspeech}. Therefore, in this work, we preprocess the data by first splitting teacher utterances into EDUs. 

Since no state-of-the-art method currently exists for automatically dividing the text into EDUs, we use the following strategy: (1) Punctuation Removal: first, we remove all punctuation from the utterances; (2) Punctuation Restoration: next, we restore the punctuation using a model trained for this task;\footnote{\url{https://huggingface.co/oliverguhr/fullstop-punctuation-multilang-large}} (3) Comparison and Segmentation: finally, we compare the original utterance with the punctuation-restored version. If the restored punctuation replaces a comma in the original text with a period, question mark, or exclamation mark, we split the utterance at that comma, thereby creating separate EDUs. By default, we also split different sentences into separate EDUs. Each EDU that resulted from the original utterance through this process inherits the original label assigned to the full utterance in {\tt MathDial} (i.e., one of the four high-level categories).

After the data is split into EDUs, the number of resulting teacher utterances in the \textbf{train} split is \textbf{5,174}. The \textbf{validation} and \textbf{test} sets are similarly segmented into EDUs and limited to \textbf{100} teacher utterances each.

\paragraph{Annotation} Using the generated tree, a \texttt{GPT-4o}-based annotation pipeline from \citet{petukhova2025fully} is applied. Since the tree's structure aligns with the hierarchical intent relationships proposed by the authors of \texttt{MathDial}, we can reasonably expect that annotation based on this tree will reflect those relationships. For instance, if the annotation using the tree assigns the label \textit{Perturbing the Question}, the original annotation should correspondingly contain \textit{Probing}, and so on. Based on this alignment, we can evaluate the annotation quality, at least in terms of consistency with the original higher-level annotations. 

\Cref{tab:metrics_annot_mathdial} presents weighted precision (P$_w$), recall (R$_w$), and F1 (F1$_w$) as well as macro F1 (F1) scores when comparing lower-level intent annotations on the training set to the original high-level teacher move categories. The low scores are expected, given that the original teacher utterances were split into EDUs while retaining the same label. As discussed earlier, different EDUs within the same utterance would often be of distinct types, which was not accounted for in the original annotation in \texttt{MathDial}.

\begin{table}[h]
    \footnotesize
    \centering
    \begin{tabular}{cccc} \hline
         $\mathbf{P_{w}}$ & $\mathbf{R_{w}}$ & $\mathbf{F1_w}$ & $\mathbf{F1}$ \\ \hline
         0.40 & 0.38 & 0.36 & 0.27 \\ \hline
    \end{tabular}
    \caption{Evaluation of 11-label annotation on the training set, comparing the new alignment with the original 4-label annotation from \texttt{MathDial}, using the annotation framework from \citet{petukhova2025fully}.}
    \label{tab:metrics_annot_mathdial}
\end{table}
\vspace{-0.25em}

Among the 5,174 teacher utterances, 1,319 remained unchanged from the original dataset, as they originally consisted of a single EDU. Annotation results for these utterances are presented in \Cref{tab:metrics_annot_mathdial_orig_utts}. While these metrics are higher than those in \Cref{tab:metrics_annot_mathdial}, they still indicate relatively poor performance.

\begin{table}[htbp]
    \footnotesize
    \centering
    \begin{tabular}{cccc} \hline
         $\mathbf{P_{w}}$ & $\mathbf{R_{w}}$ & $\mathbf{F1_w}$ & $\mathbf{F1}$ \\ \hline
         0.48 & 0.45 & 0.43 & 0.31 \\ \hline
    \end{tabular}
    \caption{Evaluation of 11-label annotation on the training set utterances that remained unchanged (i.e., originally consisted of a single EDU), comparing the new alignment with the original 4-label annotation from \texttt{MathDial}, using the annotation framework from \citet{petukhova2025fully}.}
    \label{tab:metrics_annot_mathdial_orig_utts}
\end{table}
\vspace{-0.25em}

However, a manual analysis revealed significant inconsistencies in the original annotation. Consider the following illustrative examples:
\begin{itemize}
    \item A student initially identifies 14 as the correct final answer to the task. However, during the discussion, the student incorrectly restates the final solution as \textit{10 + 10 + 4 = 24}. The teacher responds, \textit{Is that 14?} — referring back to the earlier moment when 14 was correctly identified as the expected answer (see the full dialog in \Cref{ap:mathdial_sample}). The tree-based annotation classifies this teacher utterance as {\em Seeking Self-Correction}, corresponding to {\em Probing}. However, in the original dataset, it is labeled as {\em Telling}, which we believe is not accurate.
    \item The tutor says, \textit{You need to add brackets to (8-2) and remember the order of operations.} The student responds, \textit{Yes, I understand now. The correct equation should be 6 + (8 + 8) - 2 = 22 new books.} The teacher replies, \textit{No, I said it's 8-2, not 8+8.} Although the tree-based annotation assigns this utterance to \textit{Revealing Answer} (\textit{Telling}), the original annotation labels it as \textit{Generic}, possibly reflecting a different interpretation or contextual judgment.
    \item A student states, \textit{6 + 8 + (8 - 2) = 22.} The teacher responds, \textit{Please explain how you got 22.} The tree-based annotation categorizes this utterance as {\em Asking for Explanation}, which corresponds to {\em Probing}. However, in the original dataset, it is labeled as {\em Generic}, which does not align well with the intent of the utterance.
\end{itemize}

Given the prevalence of such unclear or ambiguous cases in the original annotation of the dataset, we cannot conclude that the tree-based annotation is inaccurate. Instead, these inconsistencies in the original annotation suggest that the discrepancies in the evaluation metrics may be due, at least in part, to ambiguities in the original dataset.

\begin{figure*}[htbp]
  \centering
  \includegraphics[width=0.8\linewidth]{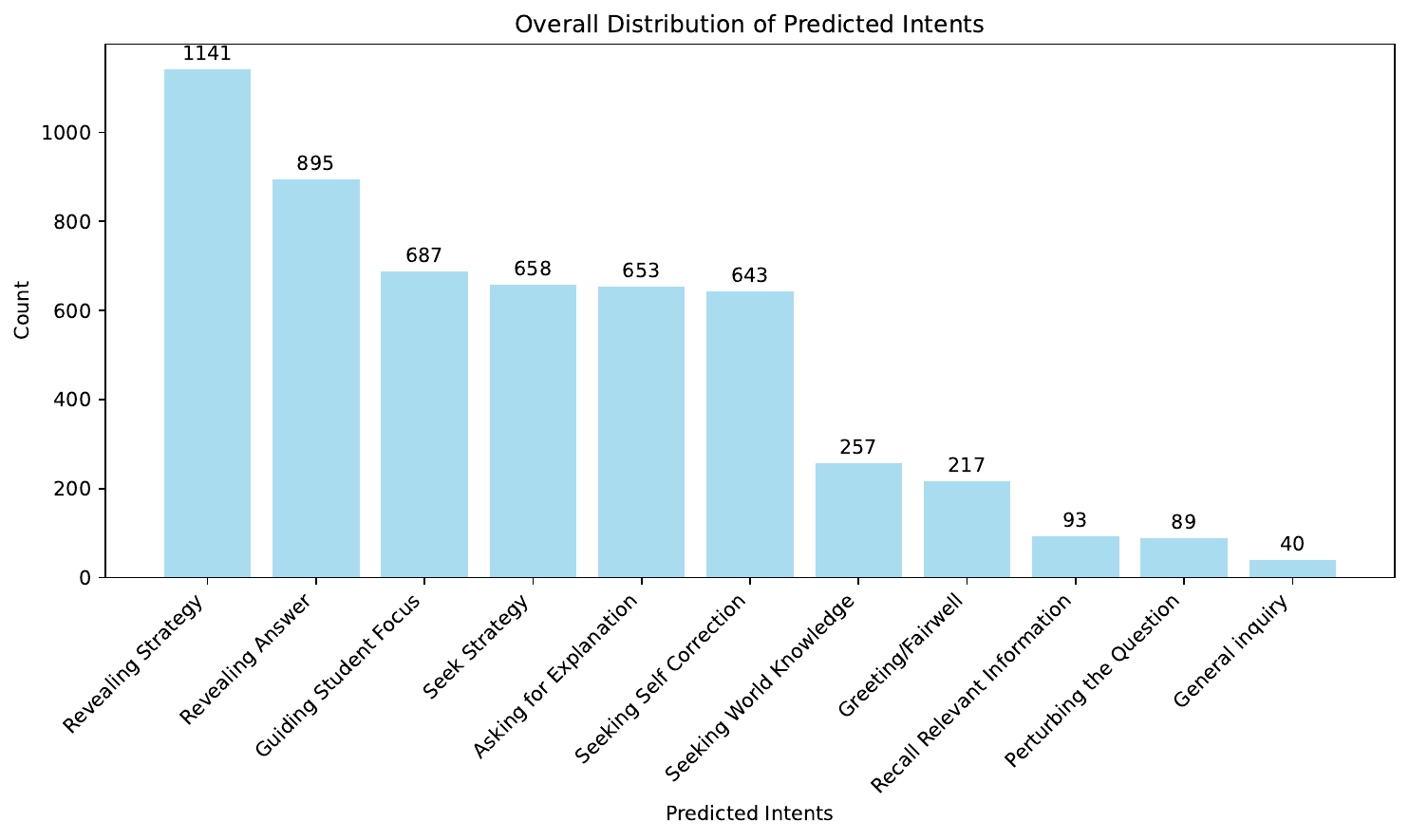}
  \caption{Overall distribution of the eleven predicted intents in the re-annotated dataset.}
  \label{fig:labels_distribution}
\end{figure*}

The distribution of the eleven predicted intents across all dataset splits (train, validation, and test) is shown in \Cref{fig:labels_distribution}.

\section{Controlled Generation}

To demonstrate the benefits of an extended taxonomy with annotations collected using the framework from \citet{petukhova2025fully}, we fine-tune an LLM to predict the next teacher utterance. The model is trained using the math task description, its gold solution, the student's solution, the dialog history, and the teacher's next utterance intent as predicted by the annotation framework.

Additionally, we fine-tune a second version of the same model using the original four-intent annotation. We then compare the performance of these two fine-tuned models with each other, as well as with the same LLM in its zero-shot setting.

\paragraph{Model} We use \texttt{Mistral-7B-Instruct} as the base model for fine-tuning, specifically its 4-bit quantized version from Hugging Face.\footnote{\url{https://huggingface.co/unsloth/mistral-7b-instruct-v0.3-bnb-4bit}} The maximum sequence length is set to 1,600. We fine-tune the model using QLoRA (Quantized Low-Rank Adaptation)~\cite{hu2022lora}, a parameter-efficient method that applies low-rank adapters with quantization to reduce memory and compute costs. We use a rank of $r=32$ and scaling factor $\alpha=32$. Fine-tuning is conducted for one epoch with a learning rate of $2\text{e}^{-5}$, batch size 8, and gradient accumulation of 4. We employ the AdamW optimizer~\citep{loshchilov2018decoupled}, linear scheduling with a warmup (0.1), weight decay of 0.1, and evaluate every 50 steps using \textsc{SacreBLEU}~\citep{post-2018-call}.

\paragraph{Data Preprocessing} We convert the annotated samples into pairs of prompts and gold outputs, where each prompt consists of an instruction, the math task, the gold solution for the task, the student's solution, the dialog history, and the intent of the following teacher utterance (which is available from the annotated data). While the intent is available as an annotation during both training and evaluation -- since we have access to the gold next teacher utterance and can classify its intent -- for real-world applications this intent would need to be predicted by a separate model as part of a controlled generation pipeline. The prompt template is shown in \Cref{ap:prompt_template_ft_mathdial}.

\paragraph{Evaluation} We conduct an automatic evaluation of generated outputs using reference-based metrics, including \textsc{chrF++} (character n-gram $F$-score)~\citep{popovic-2017-chrf}, \textsc{SacreBLEU} (a weighted geometric mean of $n$-gram precision scores), and \textsc{ROUGE-1}, \textsc{ROUGE-2}, and \textsc{ROUGE-L} (recall-oriented measures of $n$-gram overlap)~\citep{lin-2004-rouge}. In addition, we conduct a small-scale human evaluation.

\begin{table*}[htbp]
    \small
    \centering
    \begin{tabular}{lccccc} \hline
         \textbf{Configuration} & \textbf{\textsc{chrF++}} & \textbf{\textsc{SacreBLEU}} & \textbf{\textsc{Rouge-1}} & \textbf{\textsc{Rouge-2}} & \textbf{\textsc{Rouge-L}} \\ \hline
         Zero-Shot, 4 intents & 16.50 & 0.93 & 8.93 & 2.19 & 7.10 \\
         Zero-Shot, 11 intents & 17.11 &  0.73 & 8.72 & 1.95 & 6.87 \\ \hline
         Fine-Tuning, 4 intents & 16.82 & 2.67 & 17.13 & 5.61 & 15.95 \\
         Fine-Tuning, 11 intents & \textbf{18.06} & \textbf{4.59} & \textbf{20.73} & \textbf{7.39} & \textbf{19.28} \\ \hline
    \end{tabular}
    \caption{Evaluation of controlled generation on the test set from \texttt{MathDial} in a zero-shot setting and with fine-tuned \texttt{Mistral}, comparing using the original four intents from \texttt{MathDial} with eleven intents annotated using the framework from \citet{petukhova2025fully}.}
    \label{tab:metrics_generation_mathdial}
\end{table*}

\paragraph{Results} \Cref{tab:metrics_generation_mathdial} presents the generation results for both zero-shot and fine-tuning settings, comparing two annotation schemes: the original four teacher intents provided in the \texttt{MathDial} dataset and the extended set of eleven intents. As expected, the fine-tuned LLM outperforms the zero-shot baseline, and the model trained on the more fine-grained, eleven-intent annotation consistently achieves higher scores across all metrics.

\begin{figure}[htbp]
\centering
\includegraphics[width=\linewidth]{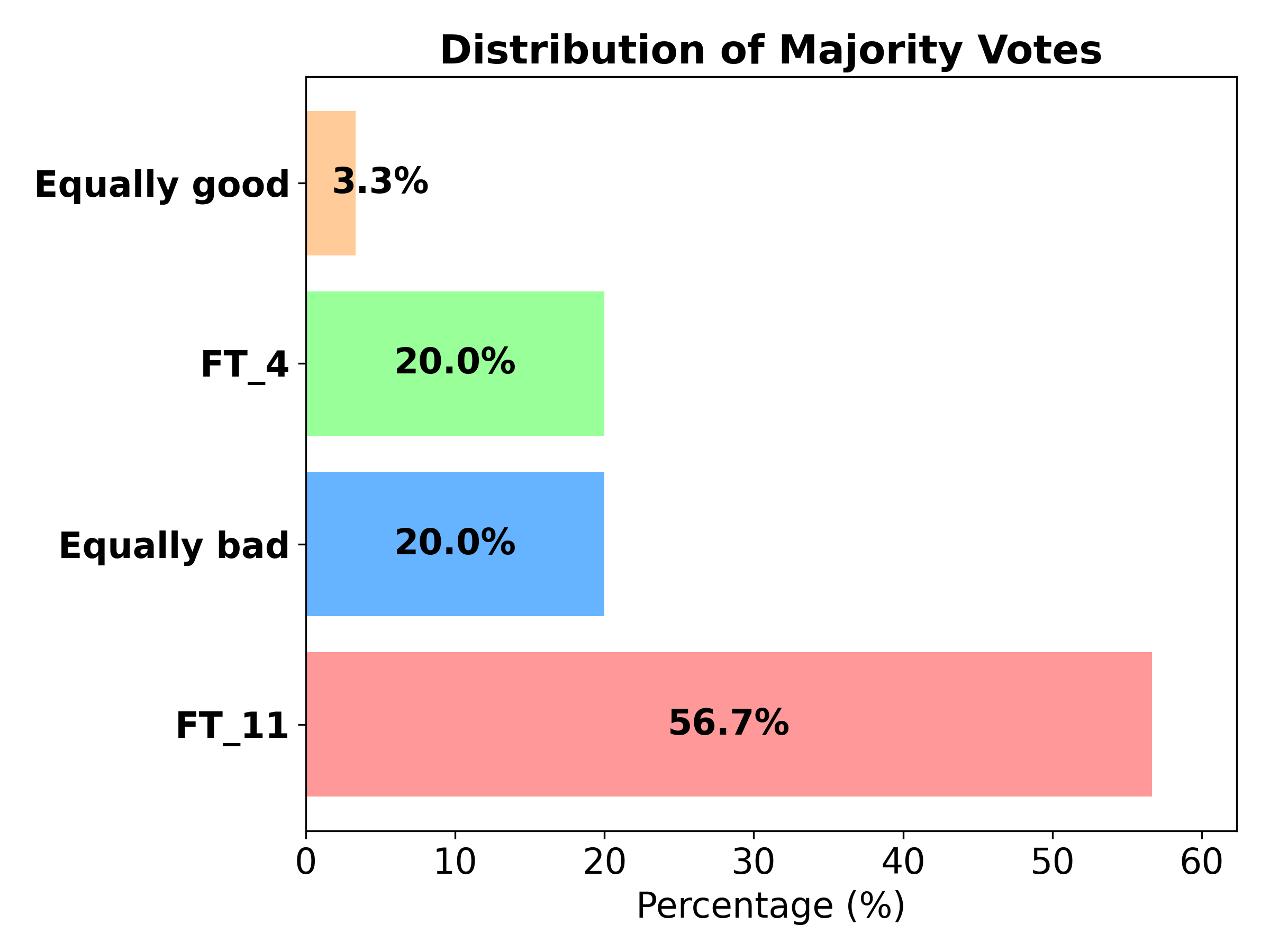}
\caption{Results of human evaluation. Four annotators were asked to choose the better response or indicate if both were equally good or bad. Final decisions were determined via a majority vote.}
\label{fig:human_eval}
\end{figure}

In addition to automated metrics, we conducted a human evaluation with four annotators, each holding at least a Master’s degree in Natural Language Processing. We randomly selected seven dialogs from the test set, resulting in 30 response pairs -- one from the model fine-tuned on four intents and one from the model fine-tuned on eleven intents. Each annotator was shown these pairs and asked to decide which response was better or whether both were equally good or poor (see \Cref{fig:human_eval}). Based on majority voting, responses from the FT-11 model were preferred in 56.7\% of cases.\footnote{There were no ties in the majority votes -- each example received a clear decision.} The inter-annotator agreement, measured using Fleiss' Kappa, is $\kappa = 0.33$, indicating fair agreement.

\begin{figure*}[!h]
  \centering
  \includegraphics[width=\linewidth]{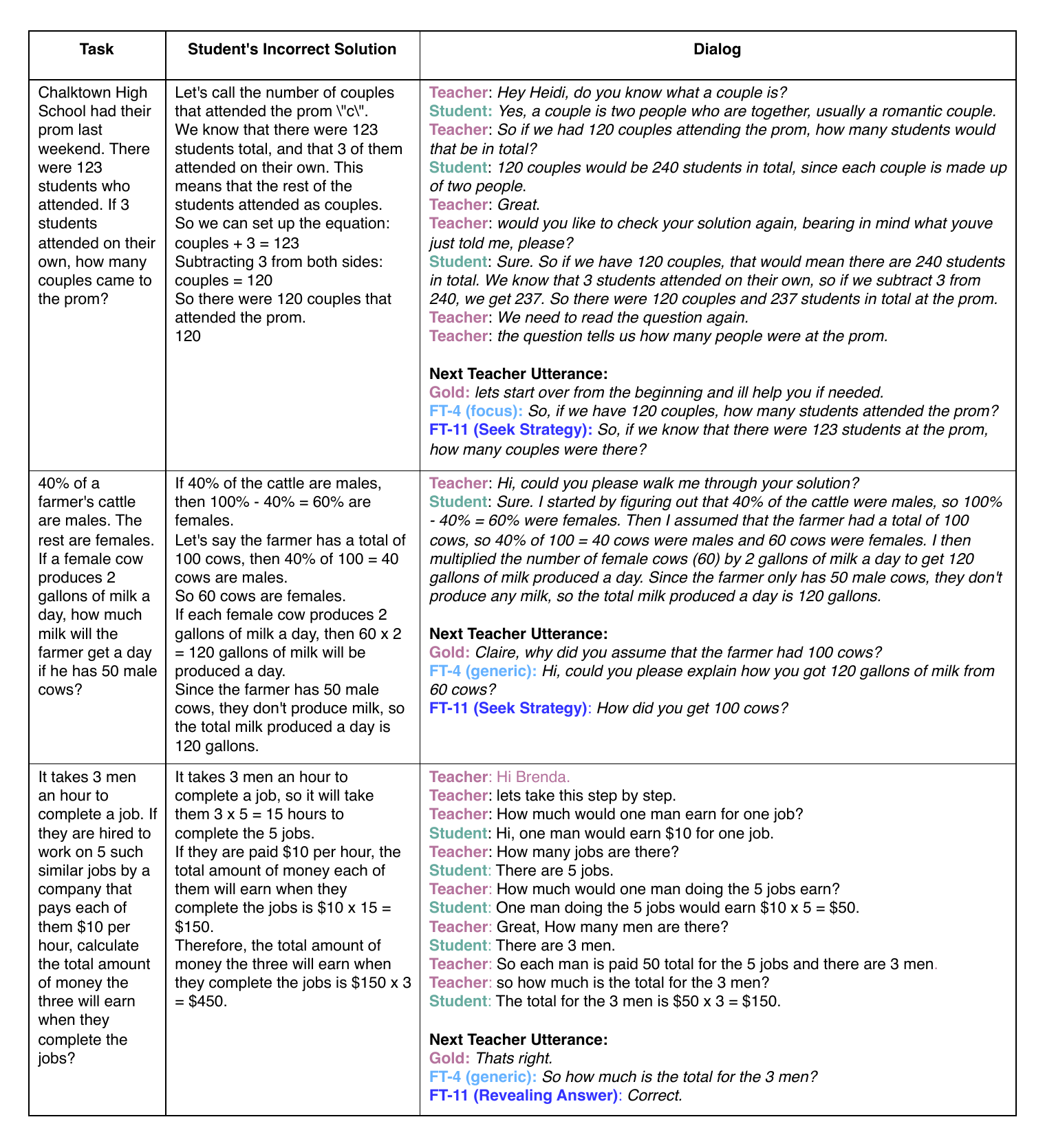}
  \caption{Examples of responses generated by a model fine-tuned on the original four-intent annotation set (FT-4) and by a model fine-tuned on data annotated with eleven intents using the framework from \citet{petukhova2025fully} (FT-11), alongside actual responses from teachers (Gold).}
  \label{fig:mathdial_generations}
\end{figure*}

Manual analysis (see \Cref{fig:mathdial_generations}) indicates that the responses generated by the model fine-tuned on eleven intents (FT-11) are superior to those from the model fine-tuned on four intents (FT-4), based on the following observations:
\begin{itemize}
    \item FT-11 consistently demonstrates a deeper understanding of conversational strategies, such as using more effective questioning techniques. For instance, when addressing the incorrect prom couples calculation (see the first example in \Cref{fig:mathdial_generations}), FT-11 explicitly prompts the student to reconsider the original conditions (\textit{So, if we know that there were 123 students at the prom, how many couples were there?}). In contrast, FT-4 merely restates the incorrect scenario (\textit{So, if we have 120 couples, how many students attended the prom?}), which is less effective in guiding the student to realize their mistake.
    \item FT-11 more directly addresses student misconceptions. In the second example in \Cref{fig:mathdial_generations}, FT-11 directly questions the student's arbitrary assumption (\textit{How did you get 100 cows?}), aligning closely with the teacher's gold standard (\textit{Claire, why did you assume that the farmer had 100 cows?}). FT-4 is less focused, requesting the student to explain calculations instead of addressing the root cause of misunderstanding.
    \item FT-11 responses tend to be concise yet relevant, prompting students to reflect critically on their reasoning rather than reiterating previous statements. For example, in the third scenario in \Cref{fig:mathdial_generations}, FT-11 succinctly acknowledges correctness (\textit{Correct.}), aligning well with the actual teacher response (\textit{That’s right.}), while FT-4 unnecessarily repeats previous questions, demonstrating less effective dialog management. 
\end{itemize}

\section{Conclusions}

In this work, we examined the impact of fine-grained annotation on controlled response generation in the \texttt{MathDial} dataset. By expanding the original taxonomy of teacher moves from four broad categories to eleven more specific intents and using the framework for automated tree creation and annotation from \citet{petukhova2025fully}, we demonstrated that this approach enhances the performance of a fine-tuned LLM in generating meaningful responses.

The results confirm that fine-tuning on a dataset with high-granularity labels leads to better alignment with expected teacher responses, outperforming both the zero-shot setting and fine-tuning on the original four-category annotation. This suggests that the specificity of intent labels is crucial for enhancing the model's ability to generate targeted and effective tutoring responses.

Furthermore, the manual analysis of the annotation process revealed inconsistencies in the original dataset, highlighting the importance of high-quality, well-structured intent annotation. The tree-based annotation framework proposed in \citet{petukhova2025fully} provides a systematic way to refine such datasets, making them more suitable for training controllable generation models.

\section*{Limitations}
Due to resource constraints, we re-annotate only a subset of the \texttt{MathDial} dataset and utilize a small language model for fine-tuning.

These limitations suggest several directions for future work: (1) re-annotating the entire dataset to enable training of higher-quality models, and (2) exploring larger open-source models for improved fine-tuning performance.

In addition, we acknowledge that our conclusions are based on the use of automated metrics and small-scale human evaluation. In order to demonstrate the real-world impact of these findings and assess the pedagogical value of the generated tutor interventions, future work should consider verifying these conclusions with actual teachers and students.

\section*{Ethical Considerations}
As this work is exploratory and the outputs of the models used in this research have not been tested with real students, we do not anticipate any significant risks associated with this work or the use of the re-annotated dataset. At the same time, we acknowledge that this work uses LLMs, and such models may present risks when applied in real-life educational scenarios, as they may generate outputs that, despite being plausible, are factually inaccurate or nonsensical, which in turn may lead to misguided decision making and propagation of biases. While we do not foresee any immediate risks associated with the research presented in our paper, if future work based on this research applies presented approaches to real-life scenarios, appropriate safeguards should be applied.

\section*{Acknowledgments}
We are grateful to Google for supporting this research through the Google Academic Research Award (GARA) 2024.

\bibliography{custom}

\appendix

\onecolumn
\section{Output of the Tree Creation Step}
\label{ap:mathdial_sample}

\begin{figure}[htbp]
  \centering
  \includegraphics[width=\linewidth]{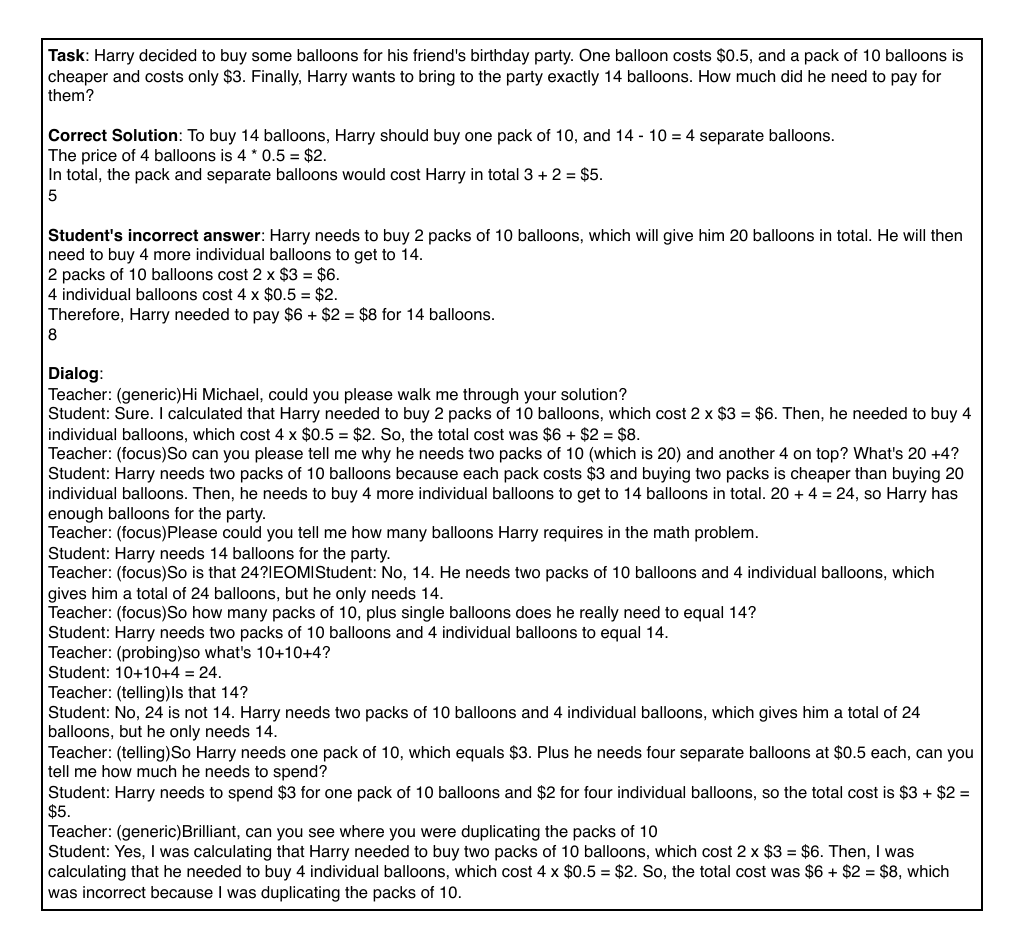}
  \caption{An example dialog from the \texttt{MathDial} dataset.}
  \label{fig:mathdial_sample}
\end{figure}

\newpage

\section{The Prompt Template used for Fine-tuning}
\label{ap:prompt_template_ft_mathdial}

\begin{figure}[htbp]
  \centering
  \includegraphics[width=\linewidth]{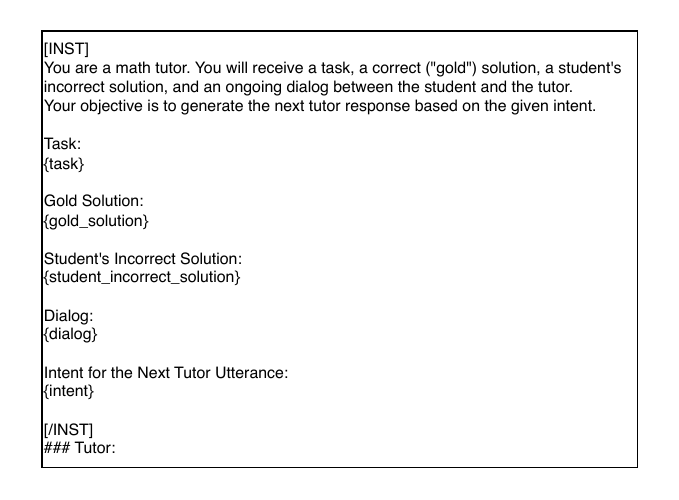}
  \caption{The prompt template used for fine-tuning on \texttt{MathDial}.}
  \label{fig:prompt_template_ft_mathdial}
\end{figure}

\end{document}